\pdfoutput=1

\documentclass[11pt]{article}

\usepackage[]{emnlp2021}

\usepackage{times}
\usepackage{latexsym}
\usepackage{tabularx}

\usepackage{graphicx}
\usepackage{multicol}
\usepackage{hyperref}

\usepackage[T1]{fontenc}

\usepackage[utf8]{inputenc}

\usepackage{microtype}

\title{MeLT: Message-Level Transformer with Masked Document Representations as Pre-Training for Stance Detection}

\author{Matthew Matero, Nikita Soni, \\ {\bf Niranjan Balasubramanian, \and H. Andrew Schwartz} \\
Department of Computer Science, Stony Brook University \\
\texttt{\{mmatero, nisoni, niranjan, has\}@cs.stonybrook.edu}}
  
\begin{document}
\maketitle
\begin{abstract}
Much of natural language processing is focused on leveraging large capacity language models, typically trained over single messages with a task of predicting one or more tokens. However, modeling human language at higher-levels of context (i.e., sequences of messages) is under-explored. In stance detection and other social media tasks where the goal is to predict an attribute of a message, we have contextual data that is loosely semantically connected by authorship. Here, we introduce Message-Level Transformer (MeLT) --  a hierarchical message-encoder pre-trained over Twitter and applied to the task of stance prediction. 
We focus on stance prediction as a task benefiting from knowing the context of the message (i.e., the sequence of previous messages). The model is trained using a variant of masked-language modeling; where instead of predicting tokens, it seeks to generate an entire masked (aggregated) message vector via reconstruction loss. We find that applying this pre-trained masked message-level transformer to the downstream task of stance detection achieves F1 performance of 67\%. 
\end{abstract}

\section{Introduction}

Generated by people, natural language data inherently spans multiple levels of analysis, from individual tokens, to documents (or messages), and to sequences of messages. While the multi-level aspect is rarely looked at beyond words-to-documents, some work has suggested benefits to modeling language as a hierarchy, such as building document representations from a collection of its sentences or a user vector given a history of their language~\cite{song-etal-2020-using, acheampong2021transformer, grail-etal-2021-globalizing, matero2019suicide, ganesan2021empirical}.

We consider stance detection, a message-level task, where the social or personal context in which the message appears (e.g., such as a person's profile) has been shown relevant to capturing the stance of the message~\cite{lynn-etal-2019-tweet, aldayel2019your}. However, such work explicitly integrated user- or social-context into the stance model, as a separate component. We ask if there is a more direct integration of user context when processing a target message. To this end, we process the target message as a part of the sequence of messages from the user. This way of using historical language from a person enables us to both model within message information (word-level) and to process the message within the author context (message-level). 


While there have been some models that take advantage of hierarchy through words and sequences of messages~\cite{lynn2020hierarchical, yu2020coupled, zhao2020pretrained} there has been little work in providing generic pre-training routines for large capacity transfer learning style models beyond the word-level. Instead, many of these hierarchical models are either applied directly to a downstream task or, if pre-trained, on an adjacent version of the downstream task. Being able to pre-train general message-level models could enable inclusion of message-level contextual information that is not easily obtainable with task-specific training that is limited in data sizes as compared to larger unlabeled corpora available for modeling at the message-level. 

In this study, we propose a hierarchical message-level transformer (MeLT) trained over a novel pre-training routine of \textit{Masked Document Modeling}\footnote{In this work a document is a single tweet (referred to as a message)}, where the goal is to encode documents in latent space using surrounding contextual documents. We then fine-tune MeLT to a stance detection dataset derived from Twitter as defined in the SemEval 2016 shared task~\cite{mohammad2016semeval}. Our contributions include: (1) introduction of a new pre-training routine for hierarchical message-level transformers\footnote{Code: \href{https://github.com/MatthewMatero/MeLT}{https://github.com/MatthewMatero/MeLT}}, (2) demonstration of efficacy of our pre-training routine for stance detection, and (3) exploratory analysis comparing model size with respect to the number of additional message-level layers and amount of user history leveraged in fine-tuning.

\section{Related Work}

Our approach is inspired by the success word-to-document level transfer learning has had since popularized by the BERT language model~\cite{devlin2018bert}. Offering the idea of a ``contextual embedding" allows models to properly disambiguate words based on their surrounding context. While other types of language models are also used, usually autoregressive based such as GPT and XLNet~\cite{brown2020language, yang2019xlnet}, many models are variants of the BERT autoencoder style~\cite{liu2019roberta,lan2019albert}.

Both ~\citet{zhang2019hibert} and ~\citet{liu2019hierarchical} use hierarchical encoder models for summarization tasks. While both models encode sentences using some surrounding context, their pre-training tasks are still that of text generation rather than latent modeling. ~\citet{yu2020coupled} encode global context in conversation threads on social media by generating a history vector (concatenated representations of each sub-thread) during the fine-tuning step and ~\citet{zhao2020pretrained} propose a capsule network to aggregate fine-tuned word representations to perform automatic stance detection. 

Stance detection is an ideal task to develop MeLT because while it is labeled at the message-level, the stance itself is presumed to be held by the author with a history of messages. Previous successful approaches to stance detection have used topic modeling, multi-task modeling via sentiment, multi-dataset training ~\cite{lin2017enhanced,li2019multi,schiller2021stance}, or user-level information ~\cite{lynn-etal-2019-tweet, aldayel2019your}. Our work builds on this by using a pre-trained transformer trained to model message representations in latent space across author histories to encode global user knowledge into individual messages. 

\section{Hierarchical Message Modeling}

Messages are made up of individual words that come together to give each other context and meaning. Comparably, a collection of messages can come together to show topics of conversation. Directly encoding the interactions of messages and their underlying words can prove beneficial when modeling language at the document or person-level. For example, processing post history of a social media user within context of their own language.

\subsection{Masked-Document Reconstruction}
We adapt the masked-language modeling (MLM) approach popularized by use in the BERT model to work for masked documents, rather than words. Namely, we introduce the \textit{masked-document modeling} task, as shown in equation \ref{eq:task_goal}, where a message sequence is ordered by created time within a user's history, some messages are selected for masking, and every message is represented as the average of their word tokens. 

\vspace*{-10pt}
\begin{equation}
    \label{eq:task_goal}
    \hat{M}_t = f(M_{t-k}, ..., masked_{t}, ..., M_{t+k}) + \epsilon
\end{equation}

Here, $\hat{M}_t$ is the reconstruction of the masked out message $M$ at step $t$ through function $f$ using the contextual messages $M_{t-k},...,M_{t-1},M_{t+1},..., M_{t+k}$ with error represented as $\epsilon$. Loss is calculated, as mean-squared-error, against the ground-truth label of the average representation of all words, $W_i$, that are present in the individual masked message shown in equations \ref{eq:loss_target} and \ref{eq:loss_def}. Thus, making the task latent space reconstruction where our model learns to encode messages by rebuilding their local representation using global context. 

\vspace*{-10pt}
\begin{equation}
    \label{eq:loss_target}
     Label = avg(W_{0},W_{1}, ... ,W_{n})
\end{equation}
\vspace*{-10pt}
\begin{equation}
    \label{eq:loss_def}
     Loss = MSE(\hat{M}_t, label)
\end{equation}

Our masking strategy follows the same rules as introduced in BERT. Specifically, a message has a 15\% chance of being selected for masking. Once selected they are then replaced with a message MASK token (80\% chance), left unchanged (10\% chance), or replaced with a random message vector (10\% chance). 

\subsection{Message-level Transformer (MeLT)}
\paragraph*{Architecture Description}
We first select a pre-trained word-level language model on which we build MeLT. This allows us to leverage models that have already shown success in many NLP tasks rather than training from scratch. 

\begin{figure}[tb]
    \centering
    \includegraphics[width=1.05\linewidth]{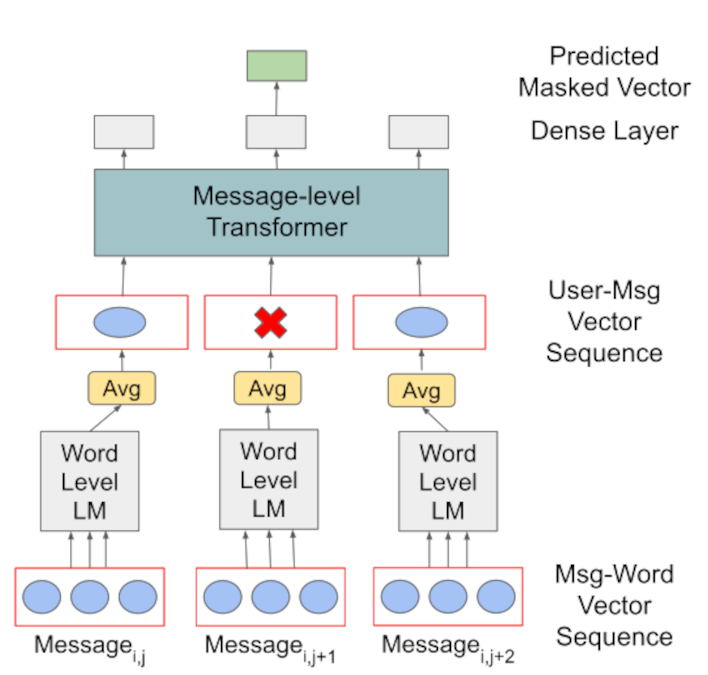}
    \caption{Pre-training architecture of our MeLT model. The bottom layer indicates a collection of a user's individual messages being processed by a word-level language model. Words within individual messages are aggregated as averages and then ordered into a sequence of 768-dimensional message-vectors per user and masking is performed, represented by a red X. Reconstruction loss is then calculated with the predicted masked vector.}
    \label{fig:pretrain_arch}
\end{figure}

After processing messages at the word-level, we average all individual word tokens within a message into a single message vector to build a sequence of message vectors and then select messages for masking. This process and architecture is highlighted in figure \ref{fig:pretrain_arch}, we refer to models using this setup as a ``Message-level Transfomer" (MeLT). Since the loss calculation as described in Eq \ref{eq:loss_def} relies on output from the word-level model itself, that portion of the model is kept frozen during pre-training. 
 
We build 2 versions of MeLT, one with 2 hierarchical layers (2L) and a 6-layer model (6L). After the last transformer layer there is a single dense linear layer which generates the final reconstructed representation of any masked out messages.

These versions of MeLT are built on top of DistilBERT (base) ~\cite{sanh2019distilbert} for the following reasons: (1) it is a smaller model (6 layers) allowing more GPU space for message-level layers and (2) while being roughly half the size of the original BERT it still offers upwards of 95\% the performance. We also explore an alternate model built-on top of DistilRoBERTa (base) to compare the utility of MeLT applied to other word-level models. 

\paragraph*{Training Instances}

For training we set the following restrictions for individual users: (1) we set a max history length of 40 for number of messages per sequence and (2) for users with more than 40 messages they are chunked and processed as separate sequences. Users with fewer than 40 total messages have message-level PAD tokens appended to their sequence. However, users that have multiple sequences will not be assigned a PAD token, if their last sequence falls short of 40 we include the amount of missing messages from their previous sequence.

\paragraph*{Dataset}
For pre-training our model we select users from publicly available tweets that were previously used for other user-level predictions, such as demographic prediction or emotion forecasting~\cite{volkova2013exploring, matero-schwartz-2020-autoregressive}. A subset of data is selected as our pre-training dataset, approximately 10 million tweets sampled from 6 thousand users, resulting in a dataset 1.3 GB in size. We use a limited dataset to highlight the utility of the pre-training routine itself and not rely on ``bigger is better" mindset. 

\begin{table}[tb]
\centering
\begin{small}
\begin{tabular}{l|c|c|c|c}
    \hline
    \textbf{Model} & \textit{F1} & \textit{Prec} & \textit{Recall} & \textit{SemEval F1} \\
    \hline
    MFC & 54 & 67 & 78 & 67 \\
    (Zarrella, 2016) & \textbf{--} & \textbf{--}& \textbf{--} & 68$\dagger$ \\
    (Zhao, 2020) & \textbf{--} & \textbf{--} & \textbf{--} & 78$\dagger$ \\
    DistilBert & 60 & 60 & 63 & 63 \\
    DistilBert + Hist & 63 & 64 & 65 & 68 \\
    (Lynn, 2019) & 66 & \textbf{--} & \textbf{--} & \textbf{--} \\
    MeLT & \textbf{67} & \textbf{68} & \textbf{67} & \textbf{73}\\
    \hline
\end{tabular}
\caption{Evaluation of various methods applied to SemEval stance detection. We report both weighted F1/Prec/Recall and Avg pos/neg F1 as defined in the original shared task. MFC is a most frequent class baseline, DistilBert and DistilBert + Hist represent an average message vector extracted from DistilBERT with or without concatenation of an average vector representing user history, respectively. MeLT is our best performing variant. \textbf{Bold} results are found significant with $p < .05$ w.r.t DistilBert + Hist using a paired t-test. ($\dagger$) indicates a model trained on the original version of the SemEval2016 dataset (4,100 total tweets) which we did not have available due to accounts or messages being deleted on twitter since release.}. 
\label{tab:table1}
\end{small}
\end{table}

\begin{figure}[tb]
    \centering
    \includegraphics[width=1.05\linewidth]{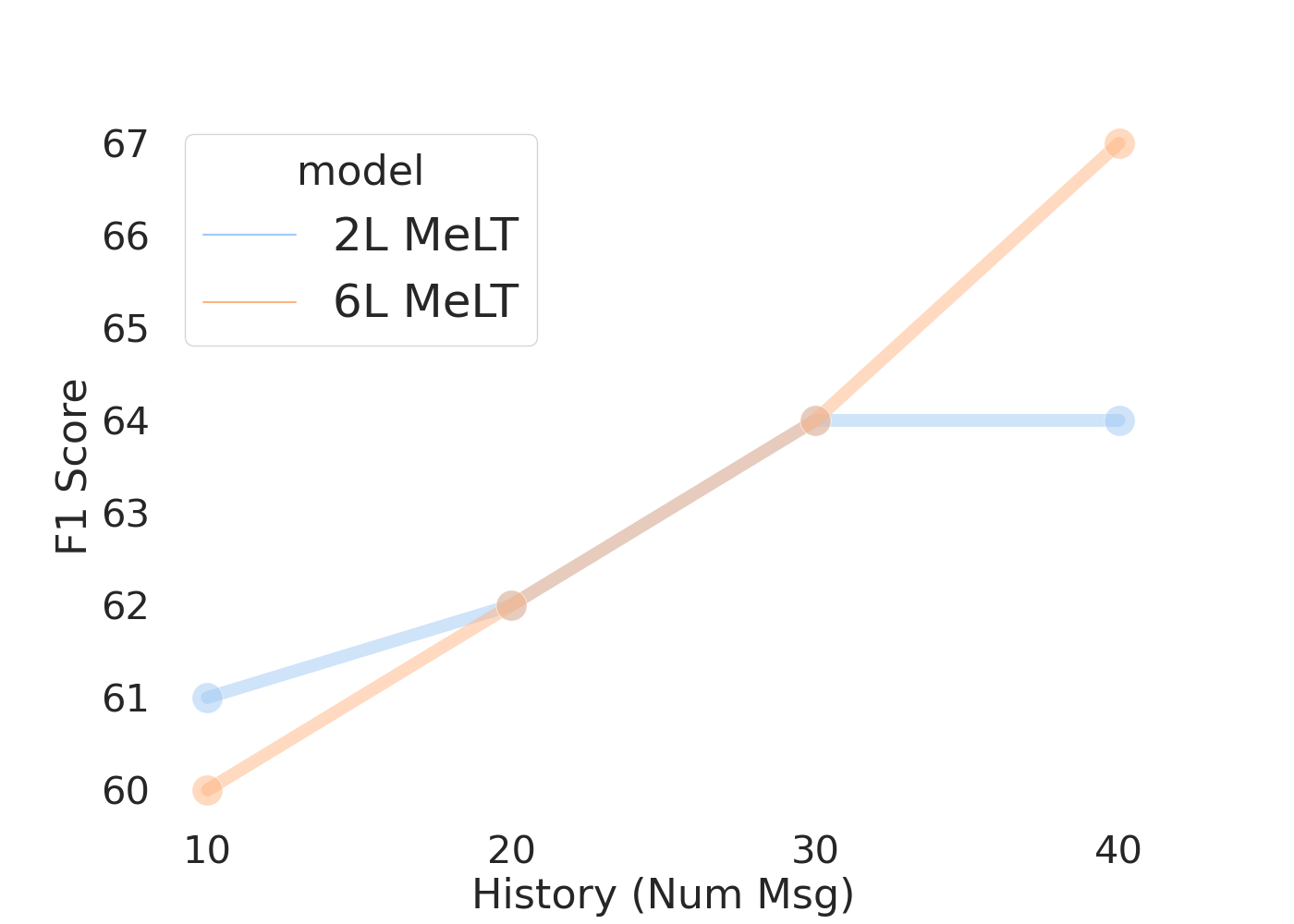}
    \caption{Average weighted-F1 performance across our models when we fine-tune using different amounts of user history. Both size MeLTs improve when more history is available, with a plateau occurring on the 2-layer model.}
    \label{fig:history_performance}
\end{figure}

\begin{table*}[tb!]
\centering
\begin{tabular}{l|c|c|c|c|c|c}
    \hline
    \textbf{Model} & \textit{Abortion} & \textit{Atheism} & \textit{Climate} & \textit{Clinton} & \textit{Feminism} & \textit{All(Avg)}\\
    \textit{Word-level Pre-train}&&&&&& \\
    \hspace{12pt}DistilBert & 60 & 66 & 70 & 58 & 46 & 60 \\ 
    \hspace{12pt}DistilBert + Hist & 64 & 62 & 70 & 64 & 54 & 63 \\ 
    \hline
    \textit{Msg-level Pre-train}&&&&&& \\
    \hspace{12pt}2L MeLT-rand & 56 & 62 & 61 & 47 & 46 & 54 \\ 
    \hspace{12pt}6L MeLT-rand & 56 & 62 & 61 & 47 & 46 & 54 \\ 
    \hspace{12pt}2L MeLT + frz word & 58 & 64 & 66 & 54 & 51 & 59 \\
    \hspace{12pt}2L MeLT + unfrz word & \textbf{66} & \textbf{67} & \textbf{74} & 58 & 59 & 65 \\
    \hspace{12pt}6L MeLT + frz word & 62 & 66 & 68 & 60 & 53 & 62 \\
    \hspace{12pt}6L MeLT + unfrz word & \textbf{66} & 66 & 71 & \textbf{67} & \textbf{63} & \textbf{67} \\
    \hline
\end{tabular}
\caption{Performance analysis on weighted F1 among all our models across each target within the SemEval dataset. MeLT-rand is our architecture applied directly to the task(no pre-train routine) and frz/unfrz word indicates whether the underlying word-level model was also updated while fine-tuning. \textbf{Bold} indicates best in column.} 
\label{tab:table2}
\end{table*}

\begin{table}[tb]
\centering
\begin{tabular}{l|c|c|c}
    \hline
    \textbf{Model} & \textit{F1} & \textit{Prec} & \textit{Rec} \\
    \textit{Word-level Pre-train}&&& \\
    \hspace{8pt} DistilRoBERTa & 59 & 55 & 57 \\
    \hspace{8pt} DistilRoBERTa + History & 61 & 68 & 66 \\
    \hline 
    \textit{Msg-level Pre-train}&&& \\
    \hspace{8pt} 2L MeLT DistilRoBERTa & 62 & 69 & 66 \\
    \hspace{8pt} 6L MeLT DistilRoBERTa & \textbf{64} & \textbf{69} & \textbf{69} \\
    \hline
\end{tabular}
\caption{Evaluation of using a different word-level model for our experiments (DistilRoBERTa). All MeLT variants are fine-tuned with the word-level model unfrozen. While we do not see this version outperform the DistilBERT variant, there are still clear benefits from using MeLT over just the word-level distil-RoBERTa. \textbf{Bold} results are found to be significant with $p < .05$ w.r.t DistilRoBERTa + History.}
\label{tab:table3}
\end{table}

\section{Stance Detection with MeLT}
We use the stance dataset available from the SemEval 2016 shared task~\cite{mohammad2016semeval}. This data includes tweets that were annotated either against, neutral, or favoring of a specific target mentioned within the tweet, across 5 distinct targets in the dataset. However, this data only includes labeled tweets from users and not any history, so we use the extended dataset from ~\citet{lynn-etal-2019-tweet}. 

During fine-tuning we keep a max history length of 40 and a temporal ordering within sequence. We apply a 2-layer feed-forward neural net with a Sigmoid activation on top of our MeLT and leave all message transformer layers unfrozen. Experiments with both frozen and unfrozen word-level layers are also explored. The message vector representation from the top transformer layer of MeLT is used as input into the fine-tuning layers.  

\section{Results}

We show a comparison of our best MeLT model against other approaches in table \ref{tab:table1}. First, we include a heuristic baseline of most-frequent-class prediction. Next, we compare against fine-tuning our word-level model of choice directly to the downstream task using 2 configurations. The first is using only the message representation, while the second is ``+ history" where we concatenate it with the average of 40 recent messages. This allows the model to have a global context within user. We also include the top participant from the shared task ~\citet{zarrella2016mitre} which uses a different F1 score as defined for the shared task, referred to here as \textit{SemEval F1}\footnote{This F1 score instead reports an average of the F-score for the positive and negative classes. Not directly accounting for neutral predictions.}. Lastly, we compare our results to the approach of  ~\citet{lynn-etal-2019-tweet}, from whom we received the extended history dataset, which uses the labeled tweet and a list of accounts the author follows. However, they only report the weighted-F1 score for their best performing model. 

We find that fine-tuning DistilBERT directly to the task of stance detection proves difficult, only scoring a modest F1. However when we include some context language from the user, an average representation of their recent language concatenated into the fine-tuning layer, there is a noticeable boost in performance highlighting that stance prediction is aided by knowing the context of the message. We find that MeLT can utilize this contextual information best and out-performs other approaches. 

Next, we break down the performance of various configurations of our models in table \ref{tab:table2} across each target. Here, we compare against a small variant of MeLT (2Layers), randomly initialized MeLTs (No pre-train)\footnote{Both MeLT-rands learn the MFC baseline}, and also experiments with frozen and unfrozen word-level models. Ultimately, we find that fine-tuning both the word and message levels simultaneously consistently proves beneficial, likely due to the word model being able to adapt to discourse on Twitter. 

We also find that the 2-layer MeLT performs competitively - in figure \ref{fig:history_performance} we show that it performs better or on-par with the large model until 40 messages of history is reached, due to the 2-layer model saturating at history of 30. Suggesting that the larger the model, the more history it can efficiently track. 

Lastly, we investigate using a different word-level model for our experiments. We choose DistilRoBERTa, for similar reasons to our original choice of DistilBERT, and apply the same techniques as done with DistilBERT shown in table \ref{tab:table3}. We find that overall each DistilRoBERTa model achieves lower F1 score than the respective DistilBERT variant. However we find that MeLT still improves over the base word-level model, suggesting that MeLT often will improve the word-level model itself but the word-level model of choice plays an important role in downstream performance. Due to this, it is likely to be beneficial to first evaluate a variety of word-level models on your downstream task and then build on top of the best one with MeLT.   


\section{Conclusion}
With a large number of tasks in NLP that rely on social media as a domain, methods which can model language as a multi-level phenomena, from words to documents to people, can offer a higher-level contextual representation of language. In this work, we presented a new hierarchical pre-training routine that, when fine-tuned to stance detection, outperforms other models utilizing both message and user-level information as well as improves results upon solely using the word-level model on which we build MeLT. We also find that during fine-tuning, it was always beneficial to unfreeze the word layers even though they had to be frozen during pre-training. MeLT can be attached to the top of a word-level language model in order to directly encode sequences of message vectors, thus allowing the modeling of historical context and leading towards a way of approaching language modeling that integrates its personal context. 

\section{Acknowledgements}
This work was supported in part by a grant from the National Institutes of Health, R01 AA028032-01 and in part by a grant from the National Science Foundation, IIS-1815358.

\bibliography{anthology,custom}
\bibliographystyle{acl_natbib}
\clearpage
\appendix

\section{Appendix}
\label{sec:appendix}

\subsection{Implementation and Hardware Details}
Pre-training of all models was performed across 3 TitanXP GPUs(12GB mem each) while fine-tuning was performed on a single TitanXP. All models were implemented using PyTorch~\cite{NEURIPS2019_9015} with the PyTorch Lightning Add-on~\cite{falcon2019pytorch}. 

During pre-training batch size was set to 100 users and fine-tuning was performed using 10. For pre-training runtime was around 2.5 hours per epoch and fine-tuning was a few minutes per epoch. MeLT 2L adds 11,621,632 trainable parameters on top of DistilBERT and MeLT 6L adds 33,677,568, as counted by summing PyTorch tensor.numel() per parameter with gradients turned on\footnote{\href{https://discuss.pytorch.org/t/how-do-i-check-the-number-of-parameters-of-a-model/4325}{https://discuss.pytorch.org/t/how-do-i-check-the-number-of-parameters-of-a-model/4325}}. All experiments (pre-training and fine-tuning) use the AdamW Optimizer~\cite{loshchilov2017decoupled} and use random seed set to 1337. Pre-training has a warm-up period of 2,000 steps. 

Pre-training is conducted over 5 epochs with checkpoints saved for the epoch that scored the lowest MSE on a holdout development set. The version of the model at that checkpoint is then used for fine-tuning to the stance dataset. 

\subsection{Hyperparams}

All hyperparameters are selected via tuning using the Optuna library~\cite{optuna_2019}.

\subsubsection{Pre-training}
The final set of hyperparameters used for the 6L MeLT model (pre-training) are as follows:

\begin{itemize}
    \setlength\itemsep{.05em}
    \item Learning Rate: 4e-3
    \item Weight Decay: 0.1
    \item Dropout: 0.1
    \item FF dim: 2048
    \item Embed dim: 768
    \item Attn Heads: 8
    \item Epochs: 5 (checkpoint at epoch 2)
    \item batch size: 100 (users)
    \item msg seq len: 40 (per user)
    \item token seq len: 50 (per message)
\end{itemize}

If any parameter is not mentioned (e.g., Adam Betas) then it uses PyTorch defaults. For pre-training 10 trials were used for parameter tuning. For pre-training only learning rate and weight decay were explored. Learning rate was searched between 5e-4 to 4e-1 and weight decay was set between 1 and 1e-4. 

\subsubsection{Fine-Tuning}

All hyperparameters were chosen based on minimizing loss over a holdout development set for each target over 50 trials. Hyper-parameters that are tuned include learning rate, weight decay, and dropout. Dropout is applied directly to output from MeLT.  Learning rate was searched between 6e-6 and 3e-3, weight decay is between 1 and 1e-4, and dropout is 0.0 to 0.05. Additionally, early stopping was also applied as a means of regularization. 

The 2-layer FFNN on top of MeLT during fine-tuning has layer 1 of dimension 768 and layer 2 of dimension 384, with Sigmoid between.

\subsection{Data}

\subsubsection{pre-training}

The pre-training dataset is comprised of 6,000 users and 9,868,429 messages. For a development set we select 3,000 users from our train set and set aside an additional 20 of their messages, to measure reconstruction loss within these sequences. 

\subsubsection{fine-tuning}

The breakdown of number of examples (labeled messages) across train/dev/test for each target in the SemEval Stance data is shown in table \ref{tab:stance_data}. In total we have 3,021 instances with a split of 1658 train, 418 dev, and 945 test across all targets. The original 2016 shared task had 4,100 instances, however due to accounts or messages being deleted over time, we were unable to replicate the complete original dataset and instead used the smaller version available from ~\citet{lynn-etal-2019-tweet}. 

\begin{table}[h]
    \centering
    \begin{tabular}{l|c|c|c}
         Target & Train & Dev & Test \\ \hline
         \hspace{6pt} Abortion & 380 & 96 & 207 \\
         \hspace{6pt} Atheism & 329 & 83 & 178 \\
         \hspace{6pt} Climate Change & 257 & 65 & 145 \\
         \hspace{6pt} Hilary Clinton & 372 & 94 & 232 \\
         \hspace{6pt} Feminism & 320 & 80 & 183\\\hline
         \hspace{6pt} Total & 1658 & 418 & 945
    \end{tabular}
    \caption{Number of examples per target in SemEval data as broken down by split of the data.}
    \label{tab:stance_data}
\end{table}

\end{document}